\documentclass{article}

\usepackage[final,nonatbib]{neurips_2023}
\usepackage[utf8]{inputenc} % allow utf-8 input
\usepackage[T1]{fontenc}    % use 8-bit T1 fonts
\usepackage{hyperref}       % hyperlinks
\usepackage{url}            % simple URL typesetting
\usepackage{booktabs}       % professional-quality tables
\usepackage{amsfonts}       % blackboard math symbols
\usepackage{nicefrac}       % compact symbols for 1/2, etc.
\usepackage{microtype}      % microtypography
\usepackage{xcolor}         % colors

\hypersetup{
    colorlinks=true,
    linkcolor=blue,
    filecolor=magenta,      
    urlcolor=teal,
}

\usepackage{subfigure}
\usepackage{multirow}
\usepackage{amsmath}
\usepackage{amssymb}
\usepackage{graphicx}
\usepackage{biblatex}
\bibliography{ref}
\usepackage{caption}
\usepackage{lipsum}

\title{WordArt Designer API: User-Driven Artistic Typography Synthesis with Large Language Models on ModelScope}

\author{
Jun-Yan He\textsuperscript{\rm{1}}\thanks{\ \ Project Leader, Alibaba Team} \quad 
Zhi-Qi Cheng\textsuperscript{\rm{2}}\thanks{\ \ Co-Lead, CMU/RCA/ICL Team} \quad 
Chenyang Li\textsuperscript{\rm{1}} \quad 
Jingdong Sun\textsuperscript{\rm{2}} \quad 
Wangmeng Xiang\textsuperscript{\rm{1}} \\
\textbf{Yusen Hu}\textsuperscript{\rm{2,5}} \quad 
\textbf{Xianhui Lin}\textsuperscript{\rm{1}} \quad 
\textbf{Xiaoyang Kang}\textsuperscript{\rm{1}} \quad 
\textbf{Zengke Jin}\textsuperscript{\rm{3,4}} \quad 
\textbf{Bin Luo}\textsuperscript{\rm{1}} \\
\textbf{Yifeng Geng}\textsuperscript{\rm{1}} \quad 
\textbf{Xuansong Xie}\textsuperscript{\rm{1}} \quad 
\textbf{Jingren Zhou}\textsuperscript{\rm{1}} \\
\textsuperscript{1}Alibaba DAMO Academy \
\textsuperscript{2}Carnegie Mellon University \
\textsuperscript{3}Zhejiang Sci-Tech University \\
\textsuperscript{4}Royal College of Art \
\textsuperscript{5}Imperial College London\
}

\begin{document}
\maketitle

\begin{figure*}[!ht]
\begin{center}
\vspace{-10mm}
\includegraphics[width=0.8\linewidth]{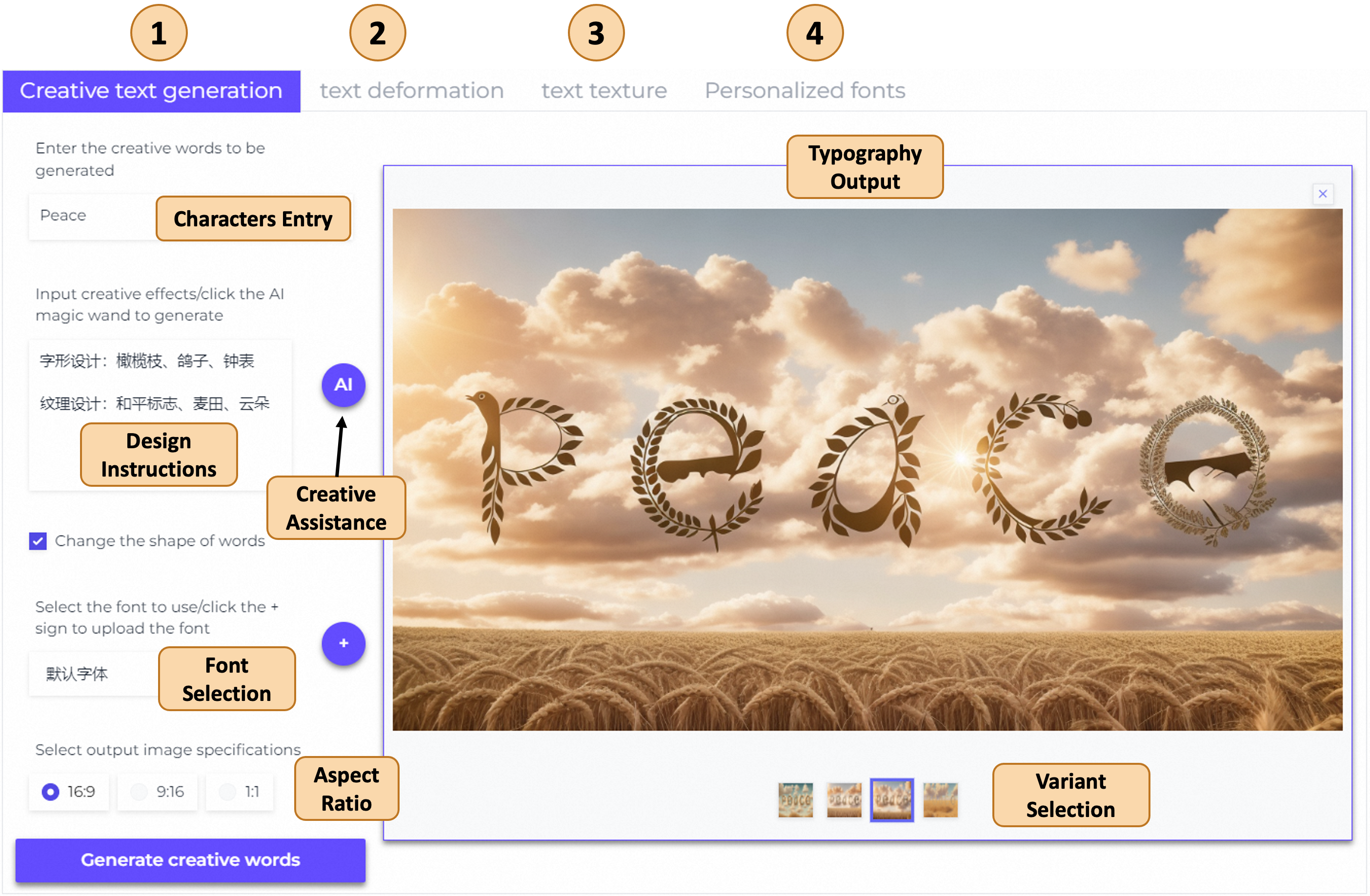}
\end{center}
\vspace{-3mm}
\captionsetup{font=small} 
\caption{Annotated screenshot showcasing our system deployed on ModelScope. Four main services are provided: (1) A "one-stop shop" that directly outputs artistic transformations based on the user input. (2) and (3) Fine-grained control of intermediate results from the SemTypo, StyTypo, and TexTypo modules. (4) Personalized font generation using handwriting. You are welcome to explore more of WordART Designer at \url{https://www.modelscope.cn/studios/WordArt/WordArt}}
\label{fig:introduction}
\vspace{-5pt}
\end{figure*}

\section{Introduction}
Typography, bridging language and design, is crucial in domains like e-commerce~\cite{cheng2016video,cheng2017video,cheng2017video2shop}, education~\cite{Vungthong2017}, and tourism~\cite{AMAR201777}. However, creating artistic typography often poses a significant challenge for those without professional training. To address this, we introduce WordArt Designer (Fig.~\ref{fig:introduction}), a user-driven framework that synthesizes artistic typography using Large Language Models (LLMs). This tool democratizes typography, making it accessible to non-experts without compromising on aesthetics or functionality.

The WordArt Designer (Fig.~\ref{fig:introduction}) initiates with an LLM module interpreting user inputs, leading to an iterative design generation process. The workflow, illustrated in Appendix~\ref{apn:framework-diagram}, begins with the LLM module interpreting user input. This generation process is iterative, in that a quality assessment feedback system automatically retries designs until at least a set number of successful transformations is produced, ensuring the creation of high-quality WordArt designs.

WordArt Designer~\cite{he2023wordart} can be employed in various areas including media, advertisement, and product design, enhancing the efficiency and effectiveness of these systems, making them more practical for everyday use. Our research not only contributes to the field of text synthesis but also opens up various practical applications.

For additional examples of typographies created by WordArt, see Appendix~\ref{apn:additional-examples}.
We also invite you to explore the capabilities of WordArt Designer further at ModelScope WordArt Studio (\url{https://www.modelscope.cn/studios/WordArt/WordArt}).

\section{Method}
\textbf{Technical Details.} We elaborate on the technical underpinnings of WordArt Designer, as introduced in our previous work \cite{he2023wordart}. The system is anchored by three key typography synthesis modules and is powered by a Large Language Model (LLM), such as GPT-3.5, to enable an interactive and user-centric design experience.

In the system's workflow, depicted in Fig. \ref{fig:framework-flow}, users commence by specifying their design requirements, which may range from conceptual themes to specific domains, such as "A cat in jewelry design." The LLM engine processes this input, converting it into structured prompts that guide the SemTypo, StyTypo, and TexTypo modules in realizing the user's artistic vision. The LLM adeptly translates free-form user inputs into precise instructions for these modules to create the desired artistic fonts. The Semantic Typography (SemTypo) module then manipulates typography based on the specified semantic concept (e.g., "Jewelry"). This process involves three key stages: (1) Character Extraction and Parameterization, utilizing FreeType \cite{david_turner_freetype_1996}, (2) Selection of Regions for Transformation, and (3) Execution of Differentiable Rasterization \cite{Li_DiffVG_SIGGraph20}. Following this, the Stylization Typography (StyTypo) module applies additional style enhancements, leveraging the Depth2Image technique from the Latent Diffusion Model \cite{Rombach_LDM_CVPR22}. It also incorporates a ResNet \cite{DBLP:conf/cvpr/HeZRS16} trained on our bespoke character dataset \cite{he2023wordart} for ranking and selecting the most successful stylizations, as further detailed below. Finally, the Texture Typography (TexTypo) module, derived from ControlNet \cite{Zhang_ControlNet_Corr23}, is responsible for producing the final textured font images.

Besides, the WordArt Designer framework has a quality assessment feedback mechanism that ensures high-quality WordArt designs. This mechanism guarantees a minimum of $K$ successful transformations through the ranking model, as shown in Fig.~\ref{fig:framework-flow}. If the system fails to meet this threshold, it automatically initiates a new iteration. The new iteration involves the LLM engine, the SemTypo and StyTypo modules, and the format directives.

\textbf{WordART Designer API on ModelScope.} The WordART Designer API serves as a dynamic tool for artistically transforming text. The process begins with users inputting text they wish to aesthetically enhance. Following this, they provide detailed specifications to guide the deformation and textural stylization, thus tailoring the design to meet specific aesthetic goals. The LLM engine plays a crucial role in this phase, generating stylization heuristics based on the textual input, which lays the groundwork for the initial design concept. During the synthesis phase, which lasts approximately one minute, the system produces four distinct design variations, offering users a spectrum of stylistic choices. The Font Deformer service, utilizing the capabilities of the SemTypo and StyTypo modules, then allows users to modify the form and style of the text, excluding textural elements. Once a desirable stylization is achieved, users can employ the Font Texturizer to apply intricate textural (graphical) details to the stylized image.

\section{Evaluation}
Since its launch, WordART Designer's integration within ModelScope has garnered significant attention, evidenced by 146,714 visits. This substantial engagement has provided a wealth of user feedback, crucial for the ongoing refinement of our service. In response to this feedback, we are considering several enhancements for the near future, including adjustable character spacing, selective background removal, and the capability for direct image exports.

Despite these areas for improvement, the ability of WordART Designer to produce rich and aesthetically appealing typographies has been widely acknowledged. The system is currently implemented on production platforms, notably on ModelScope Studio (\url{https://www.modelscope.cn/studios/WordArt/WordArt}), and is utilized internally by various corporations, including Alibaba Group. This widespread adoption underscores the practical utility and appeal of the tool in professional settings. For a more comprehensive understanding of WordART Designer's versatility and the range of textures it can generate (e.g., organic versus metallic for contexts like food and jewelry), readers are encouraged to refer to Appendix \ref{apn:additional-examples}. These examples showcase the adaptability of WordART Designer in creating context-specific typographies, further illustrating the system's capabilities.

% \section{Ethical Implications}
% Potential ethical concerns include perpetuating cultural stereotypes due to the use of certain imagery or symbols in the process of artistic transformations, or introducing bias against under-represented cultures. Another issue could be the potential inclusion of copyrighted graphics. Users need to pay attention to these issues to ensure responsible and respectful use of the system.

\section{Ethical Implications}
The deployment of WordART Designer raises important ethical considerations. 1)~A primary concern is the risk of perpetuating cultural stereotypes, as the tool might favor dominant narratives, potentially leading to the marginalization of under-represented cultures. Ensuring diversity in the dataset and algorithmic checks for bias is crucial to mitigate this issue. 2)~Another concern is the inadvertent use of copyrighted material in designs. This necessitates the integration of copyright detection mechanisms and clear user guidelines to prevent intellectual property infringement. 3)~Additionally, the impact of AI on traditional artistic roles must be considered. While enhancing design capabilities, AI tools like WordART Designer could challenge the value of human artistry, prompting a reevaluation of AI's role in creative industries. 4)~Lastly, user privacy and data security are essential, especially given the personal nature of the designs. Adhering to strict data protection standards is imperative to ensure user trust and system integrity. Addressing these ethical concerns requires ongoing efforts, including user education, ethical guideline updates, and collaboration with diverse stakeholders to enable responsible and respectful use of WordART Designer.

\printbibliography

@String(CVPR= {IEEE Conf. Comput. Vis. Pattern Recog.})

@String(SIGGRAPH= {ACM SIGGRAPH})

@Inproceedings{Rombach_LDM_CVPR22,
    author    = {Rombach, Robin and Blattmann, Andreas and Lorenz, Dominik and Esser, Patrick and Ommer, Bj\"orn},
    title     = {High-Resolution Image Synthesis With Latent Diffusion Models},
    booktitle = {CVPR},
    year      = {2022},
    pages     = {10684-10695}
}

@article{Li_DiffVG_SIGGraph20,
  author       = {Tzu{-}Mao Li and
                  Michal Luk{\'{a}}c and
                  Micha{\"{e}}l Gharbi and
                  Jonathan Ragan{-}Kelley},
  title        = {Differentiable vector graphics rasterization for editing and learning},
  journal      = {SIGGRAPH},
  volume       = {39},
  number       = {6},
  pages        = {193:1--193:15},
  year         = {2020},
}

@article{Zhang_ControlNet_Corr23,
  author       = {Lvmin Zhang and
                  Maneesh Agrawala},
  title        = {Adding Conditional Control to Text-to-Image Diffusion Models},
  journal      = {arXiv preprint},
  volume       = {abs/2302.05543},
  year         = {2023}
}

@inproceedings{DBLP:conf/cvpr/HeZRS16,
  author       = {Kaiming He and
                  Xiangyu Zhang and
                  Shaoqing Ren and
                  Jian Sun},
  title        = {Deep Residual Learning for Image Recognition},
  booktitle    = {2016 {IEEE} Conference on Computer Vision and Pattern Recognition,
                  {CVPR} 2016, Las Vegas, NV, USA, June 27-30, 2016},
  pages        = {770--778},
  year         = {2016}
}

@software{david_turner_freetype_1996,
  title = {{{FreeType}} 2},
  author = {{David Turner} and {Robert Wilhelm} and {Werner Lemberg}},
  year = {1996},
  url = {https://freetype.org/index.html},
  urldate = {2023-07-24},
  organization = {{FreeType}}
}

@article{AMAR201777,
title = {Typography in destination advertising: An exploratory study and research perspectives},
journal = {Tourism Management},
volume = {63},
pages = {77-86},
year = {2017},
issn = {0261-5177},
doi = {https://doi.org/10.1016/j.tourman.2017.06.002},
url = {https://www.sciencedirect.com/science/article/pii/S0261517717301243},
author = {Jennifer Amar and Olivier Droulers and Patrick Legohérel},
}

@article{Vungthong2017,
author = {Vungthong, Sompatu and Djonov, Emilia and Torr, Jane},
title = {Images as a Resource for Supporting Vocabulary Learning: A Multimodal Analysis of Thai EFL Tablet Apps for Primary School Children},
journal = {TESOL Quarterly},
volume = {51},
number = {1},
pages = {32-58},
doi = {https://doi.org/10.1002/tesq.274},
url = {https://onlinelibrary.wiley.com/doi/abs/10.1002/tesq.274},
eprint = {https://onlinelibrary.wiley.com/doi/pdf/10.1002/tesq.274},
year = {2017}
}

@inproceedings{cheng2017video2shop,
  title={Video2shop: Exact matching clothes in videos to online shopping images},
  author={Cheng, Zhi-Qi and Wu, Xiao and Liu, Yang and Hua, Xian-Sheng},
  booktitle={Proceedings of the IEEE conference on computer vision and pattern recognition},
  pages={4048--4056},
  year={2017}
}

@article{cheng2017video,
  title={Video ecommerce++: Toward large scale online video advertising},
  author={Cheng, Zhi-Qi and Wu, Xiao and Liu, Yang and Hua, Xian-Sheng},
  journal={IEEE transactions on multimedia},
  volume={19},
  number={6},
  pages={1170--1183},
  year={2017},
  publisher={IEEE}
}

@inproceedings{cheng2016video,
  title={Video ecommerce: Towards online video advertising},
  author={Cheng, Zhi-Qi and Liu, Yang and Wu, Xiao and Hua, Xian-Sheng},
  booktitle={Proceedings of the 24th ACM international conference on Multimedia},
  pages={1365--1374},
  year={2016}
}

@inproceedings{he2023wordart,
  title={WordArt Designer: User-Driven Artistic Typography Synthesis using Large Language Models},
  author={He, Jun-Yan and Cheng, Zhi-Qi and Li, Chenyang and Sun, Jingdong and Xiang, Wangmeng and Lin, Xianhui and Kang, Xiaoyang and Jin, Zengke and Hu, Yusen and Luo, Bin and others},
  booktitle={Proceedings of the 2023 Conference on Empirical Methods in Natural Language Processing},
  year={2023}
}

\newpage

\appendix

\section{System Framework Diagram}
\label{apn:framework-diagram}

\begin{figure}[ht]
    \centering
    \includegraphics[width=\textwidth]{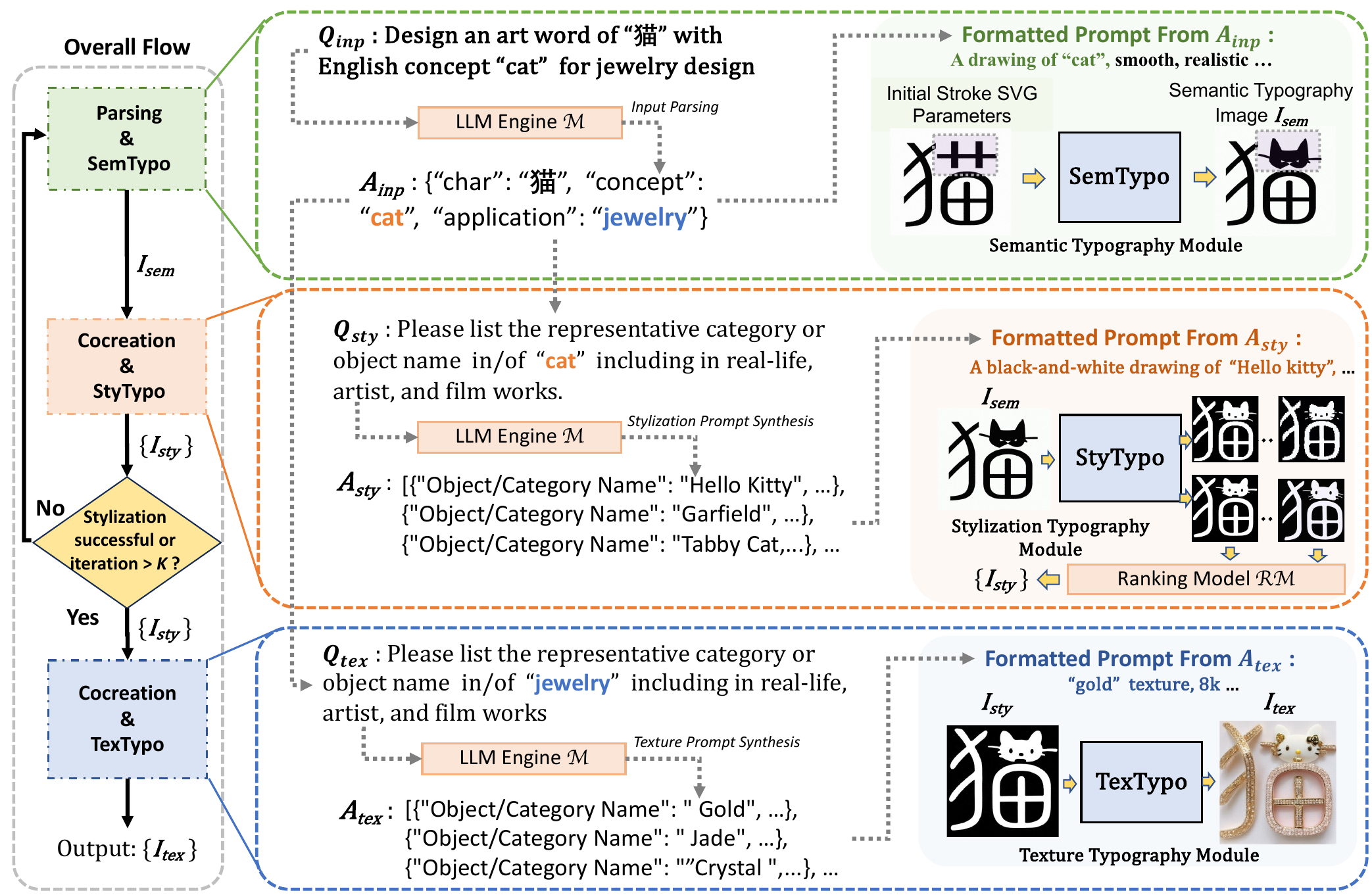}
    \captionsetup{font=small} 
    \caption{WordART Designer's various components and how they interact with each other.}
    \label{fig:framework-flow}
\end{figure}

\section{Additional Examples of Fonts Generated by WordART Designer}
\label{apn:additional-examples}

\begin{figure}[ht]
    \centering
    \includegraphics[width=0.9\textwidth]{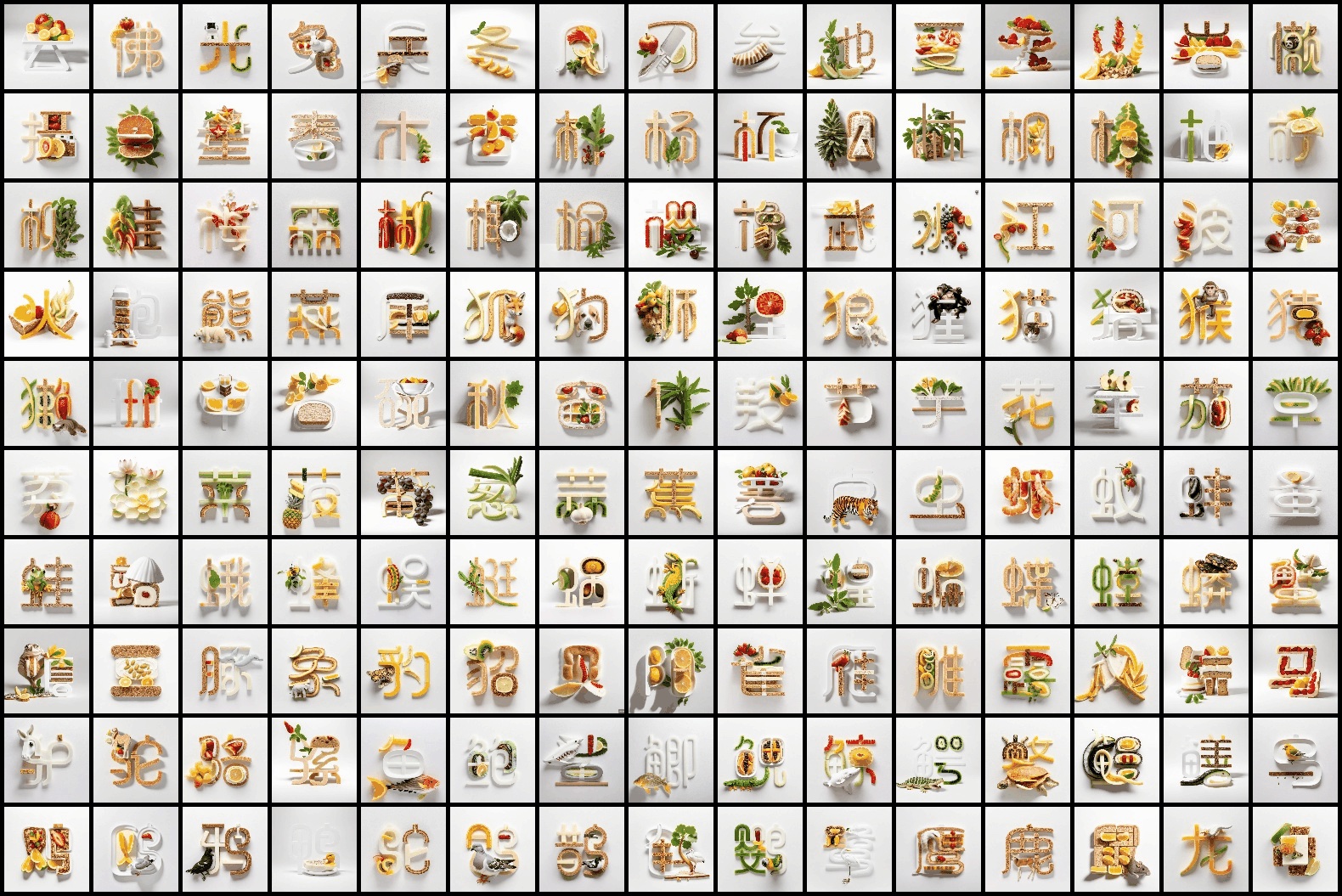}
    \captionsetup{font=small} 
    \caption{Typographies targeting fruits and foods.}
    \label{fig:fruits_example}
\end{figure}

\begin{figure}[ht]
    \centering
    \includegraphics[width=0.9\textwidth]{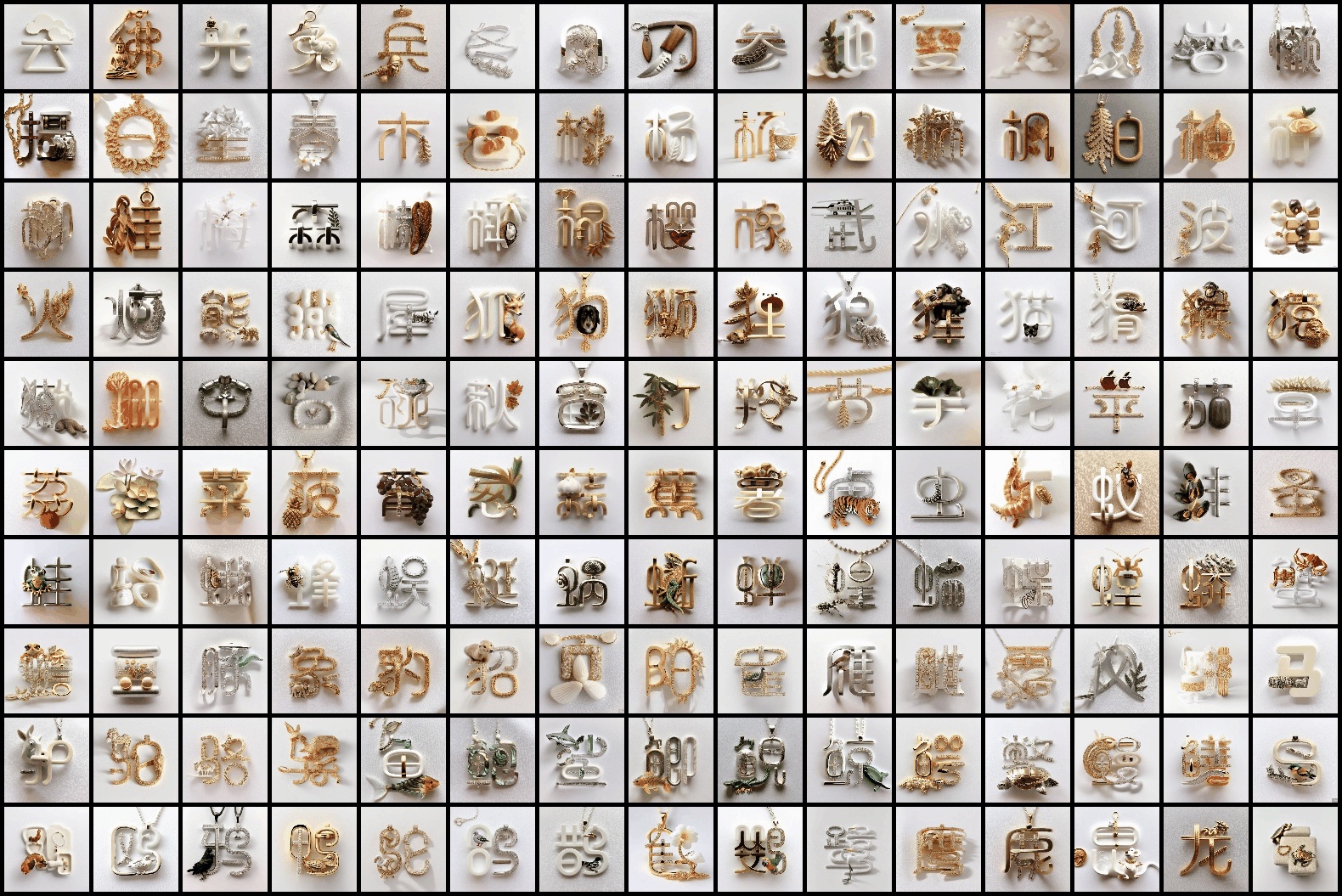}
    \captionsetup{font=small} 
    \caption{Typographies targeting jewelry.}
    \label{fig:jewelry_example}
\end{figure}

\begin{figure}[ht]
    \centering
    \includegraphics[width=0.9\textwidth]{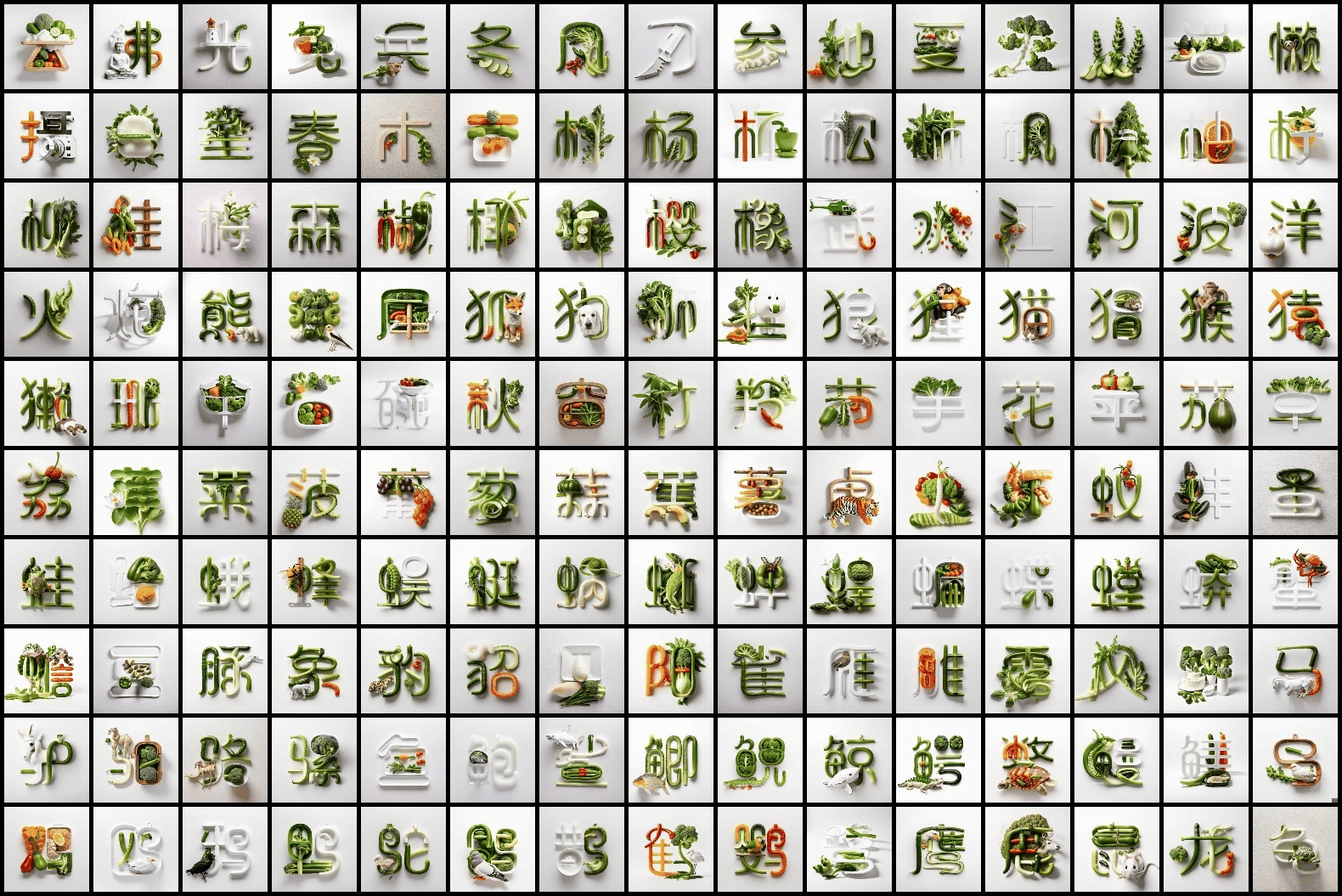}
    \captionsetup{font=small} 
    \caption{Typographies targeting plants and vegetables.}
    \label{fig:plants_example}
\end{figure}

\end{document}